\documentclass[conference]{IEEEtran}
\usepackage{times}

\usepackage{bm} 
\usepackage[numbers]{natbib}
\usepackage{multicol}
\usepackage[bookmarks=true]{hyperref}
\usepackage{amsfonts}
\usepackage{algorithm}
\usepackage{algpseudocode}
\usepackage{amsmath}
\usepackage{subcaption}
\usepackage{graphicx}
\usepackage{multicol}
\usepackage{multirow}

\DeclareMathOperator*{\argmin}{\arg\!\min}

\algnewcommand{\Inputs}[1]{%
  \State \textbf{inputs:}
  \Statex \hspace*{\algorithmicindent}\parbox[t]{.8\linewidth}{\raggedright #1}
}

\algnewcommand{\Initialize}[1]{%
  \State \textbf{initialize:}
  \Statex \hspace*{\algorithmicindent}\parbox[t]{.8\linewidth}{\raggedright #1}
}

\begin{document}

% paper title
\title{Highly Parallelized Data-driven MPC for \\ Minimal Intervention Shared Control}

\author{\authorblockN{Alexander Broad\authorrefmark{1}\authorrefmark{3},
Todd Murphey\authorrefmark{2},
Brenna Argall\authorrefmark{1}\authorrefmark{2}\authorrefmark{3}}
\authorblockA{\authorrefmark{1}Department of Computer Science}
\authorblockA{\authorrefmark{2}Department of Mechanical Engineering\\
Northwestern University, Evanston, IL 60208\\
Email: alex.broad@u.northwestern.edu}
\authorblockA{\authorrefmark{3}Shirley Ryan AbilityLab, Chicago, IL 60611}}

\maketitle

\begin{abstract}
We present a shared control paradigm that improves a user's ability to operate complex, dynamic systems in potentially dangerous environments without \textit{a priori} knowledge of the user's objective.  In this paradigm, the role of the autonomous partner is to improve the general safety of the system without constraining the user's ability to achieve unspecified behaviors.  Our approach relies on a data-driven, model-based representation of the joint human-machine system to evaluate, in parallel, a significant number of potential inputs that the user may wish to provide.  These samples are used to (1) predict the safety of the system over a receding horizon, and (2) minimize the influence of the autonomous partner.  The resulting shared control algorithm maximizes the authority allocated to the human partner to improve their sense of agency, while improving safety.  We evaluate the efficacy of our shared control algorithm with a human subjects study (n=20) conducted in two simulated environments: a balance bot and a race car. During the experiment, users are free to operate each system however they would like (i.e., there is no specified task) and are only asked to try to avoid unsafe regions of the state space.  Using modern computational resources (i.e., GPUs) our approach is able to consider more than 10,000 potential trajectories at each time step in a control loop running at 100Hz for the balance bot and 60Hz for the race car.  The results of the study show that our shared control paradigm improves system safety without knowledge of the user's goal, while maintaining high-levels of user satisfaction and low-levels of frustration.  Our code is available online at \url{https://github.com/asbroad/mpmi_shared_control}.
\end{abstract}

\IEEEpeerreviewmaketitle

\section{Introduction}
\label{sec-introduction}

Shared control is a paradigm that incorporates an autonomous partner into the control loop of a robotic system to help a human partner achieve tasks they would otherwise be unable to on their own~\cite{abbink2018topology}.  This approach offers an alternative to fully autonomous robotic systems, and can be used to extend the efficacy of modern robots in human-oriented domains (e.g., surgery, assistance and rehabilitation, search and rescue, etc$\dots$) through collaboration.  In these domains the two partners often need to communicate frequently and may even be co-located.  The most important consideration of the autonomous partner is therefore the safety of the joint system, as unsafe behavior can lead to injury to the human operator.  However, there are other features that may be equally important when we consider both the range of behaviors the human partner may wish to perform, and the user's acceptance of the assistance provided the autonomous partner.

In the majority of related work, the decision of which partner should be in control at a given time is made based on (1) the safety of the system and (2) an evaluation of who would provide better input to achieve a perceived, or known, task goal~\cite{music2017control}.  These task-specific systems can frustrate the human partner when the user's intended goal is difficult to predict, and they fail out-right when there is no desired trajectory, task or goal the user is trying to achieve.  Consider, for example, a person operating a lower-limb exoskeleton for rehabilitation purposes.  In this scenario, task-level and performance-based metrics (e.g., the amount of area covered during a search mission), are not important in determining which partner should be in control at a given time, as there is often no explicit notion of a goal (i.e., the human operator may simply want to wander around aimlessly).  Instead, the person's instantaneous desires are often the most relevant feature, conditioned  on the general safety of the system.  

The question we consider in this work then is, how does one define safety constraints that can enhance a user's ability to operate complex, dynamic machines without artificially constraining their capacity to achieve unspecified behaviors?  We address this gap in the literature with a task-agnostic control allocation strategy that balances the human's desires to achieve a wide range of possible behaviors, while simultaneously improving the safety of the joint system.  Our shared control paradigm therefore adheres to the following three ideals:
\begin{enumerate}
\item safety is paramount,
\item the user has no explicit task goal, and
\item the autonomy should exert as little influence as possible.
\end{enumerate}

In other words, the goal of our shared control paradigm is to allow the user to do whatever they would like, so long as the safety of the joint system is satisfied.  Additionally, when the autonomous partner does intervene, it should only minimally modify the user's input (a.k.a., the minimal intervention principal~\cite{anderson2012constraint}).  By adhering to these ideals, we hope to increase the influence of the human partner and consequently improve their acceptance of the assistance provided by the autonomy.

In this work, we develop a shared control algorithm that uses a highly parallelizable sampling-based model predictive control (MPC) algorithm to generate the autonomous partner's policy.  By sampling densely at uniform over the input space, we can evaluate a large set of potential actions that the user may wish to take, without \textit{a priori} knowledge of a specific goal.  We then use ideas from model-based reinforcement learning and model predictive control to generate predicted trajectories (or imagined rollouts) that represent the configuration (and safety) of the robot over a receding horizon.  Conditioned on the predicted safety of the robotic system, we iteratively select the sampled action that most closely matches the human partner's input, allowing the user to more safely move around the environment without adhering to a single objective.  Our approach relies on a representation of the human-machine system that is valid both with, and without, a known analytical model.  We focus on the latter case and therefore learn a model of the joint system offline from data.  We evaluate the efficacy of our approach with a human subjects study in two simulated environments.  Additionally, we provide an open-source, scalable implementation of our algorithm in both environments that uses a GPU for real-time interaction.

The main contributions of this work are therefore:
\begin{itemize}
\item A highly parallelizable sampling-based model predictive control algorithm for autonomous policy generation.
\item A predictive notion of safety that can evaluate the impact of a current action over a receding horizon.
\item A human-motivated cost function that only considers the instantaneous desires of the human-in-the-loop and requires no knowledge of an explicit goal.
\item Results of a human subjects study to evaluate the efficacy of, and user experience with, our shared control paradigm.
\item A GPU-implementation for real-time control of two simulated systems.
\end{itemize}

In Section~\ref{sec-background-and-related-work} we provide background information and related work.  In Section~\ref{sec-approach} we detail the theoretically exact solution to our problem.  In Section~\ref{sec-algorithmi-description} we describe our approximation, why it scales well to the majority of devices in our target domain (human-centered robotics), and provide analytical bounds on the sub-optimality of the applied control with respect to the human's desired motion.  In Section~\ref{sec-experimental-evaluation} we describe the experiment we use to validate our shared control paradigm through a human subjects study consisting of 20 participants in two simulated environments.  We also detail pertinent metrics to analyze the human partner's control skill and style, which are easily computable due to our sampling-based approach.  In Section~\ref{sec-results} we present the study results which we discuss in Section~\ref{sec-discussion}.  We conclude in Section~\ref{sec-conclusion}.

\section{Background and Related Work}
\label{sec-background-and-related-work}

In this section we discuss background and related literature in both shared and fully autonomous control, with a particular focus on human-oriented domains. 

\subsection{Shared Control}
\label{sub-sec-sc}

The majority of the shared control literature focuses on assisting a human operator when a desired task is known \textit{a priori}~\cite{kim2006continuous, carlson2008human} or predicted based on a model of the operator's intent~\cite{dragan2013policy}.  For this reason, task success is often the primary metric of concern in analyzing shared control systems, while the user's desires relating to \textit{how} a motion is achieved are often disregarded.  Related literature that addresses this same limitation in shared control includes techniques that rely on model-free control policies~\cite{reddy2018shared}, POMDPs~\cite{javdani2018shared}, and control barrier functions~\cite{broad2018operation}.  In some application domains there is a welcome trade-off between achieving the desired high-level goal and intervention from the autonomous partner (e.g., with ground vehicles on a roadway where a large amount of structure is enforced on the motion of the dynamic system).  In more human-centered domains, such as assistive and rehabilitation robotics, there is often significantly less structure imposed on the motion of the machine and there may be no explicit goal.  For this reason, the same trade-off in task success and autonomous intervention is not consistently accepted by users~\cite{erdogan2017prediction}.  In these domains, it is instead of utmost importance that the user retains a sense of personal agency and a feeling of control over the mechanical device.

Despite these differences, the most closely related work to our own (from a methodological standpoint) can be found in the semi-autonomous vehicle literature.  The semi-autonomous vehicle paradigm is distinct from the concept of a fully autonomous self-driving cars as the autonomy's goal is not to take full control of the vehicle, but instead to act as a guardian or intelligent co-pilot~\cite{anderson2013intelligent}, intervening on the person's control when deemed necessary to ensure safety.  For this reason, there is a growing line of work that follows the minimum intervention principle in the semi-autonomous vehicle domain~\cite{leung2018infusing}.  For example, Schwarting et al.~\cite{schwarting2017parallel, schwarting2017safe} describe a parallel autonomy framework that develops control trajectories for semi-autonomous vehicles that minimize deviation from user-input and achieve task-specific metrics like road following and contour tracking.  Anderson et al.~\cite{anderson2012constraint, anderson2014experimental} describe a geometric, homotopy-based algorithm for computing \textit{free space} in the environment.  The human operator is then allocated full control of the dynamic system so long as their input will not violate the constraints defined by the geometric constructs.  In this work we provide an alternative method (prediction) of computing safe space in the environment that is simpler (e.g., there is no need to integrate constraints defined by potentially complex geometries into the optimization problem), and acts over the entire control space (i.e., instead of only the steering angle~\cite{anderson2012constraint, carlson2008human, erlien2016shared}).  Additionally, all prior work in this area assumes \textit{a priori} knowledge of the system dynamics whereas our technique extends to models learned from data.  

\subsection{Sampling-based and Stochastic Optimal Control}
\label{sub-sec-soc-sbc}

From a control-theoretic standpoint, related work includes shared control algorithms that build on ideas from sampling-based optimal control.  For example, Carlson et al. have explored the idea of controlling a wheelchair using \textit{safe mini-trajectories}~\cite{carlson2008human, carlson2012collaborative}, however, this work again relies on \textit{a priori} knowledge of the human operator's goal (or predicted goals based on sensor information).  Relatedly, Shia et al.~\cite{shia2014semiautonomous} use a probabilistic model of the user's potential future inputs to achieve a pre-specified task, instead of allowing the operator the freedom to move however they would like at each instant.

Our approach is also related to fully autonomous control solutions that rely on sampling-based and stochastic optimal control methods.  For example, Lavalle et al. propose Rapidly-expanding Random Trees (RRTs)~\cite{lavalle2001randomized}, which develop random trajectories through the state space that are achievable as they are constrained by the system dynamics.  More recently, Kousik et al. extend this idea through a model predictive control algorithm that is based on the notion of a forward reachability set~\cite{kousik2017safe, kousik2018bridging} to ensure safety.  Finally, Williams et al. have proposed Model Predictive Path Integral (MPPI)~\cite{theodorou2010generalized, williams2017model} control as a method of solving the optimization problem through path integrals.  These ideas build on similar theory to our own (approximating an optimal solution from a nominally infinite set of trajectories), however, they are again all standardly defined in relation to a specified start and goal configuration.  We instead consider an approximation to the infinite set of trajectories that stem from a single point and extend in \textit{all} directions for a given time-horizon.

\begin{figure*}[t]
  \centering
  \includegraphics[width=0.80\hsize]{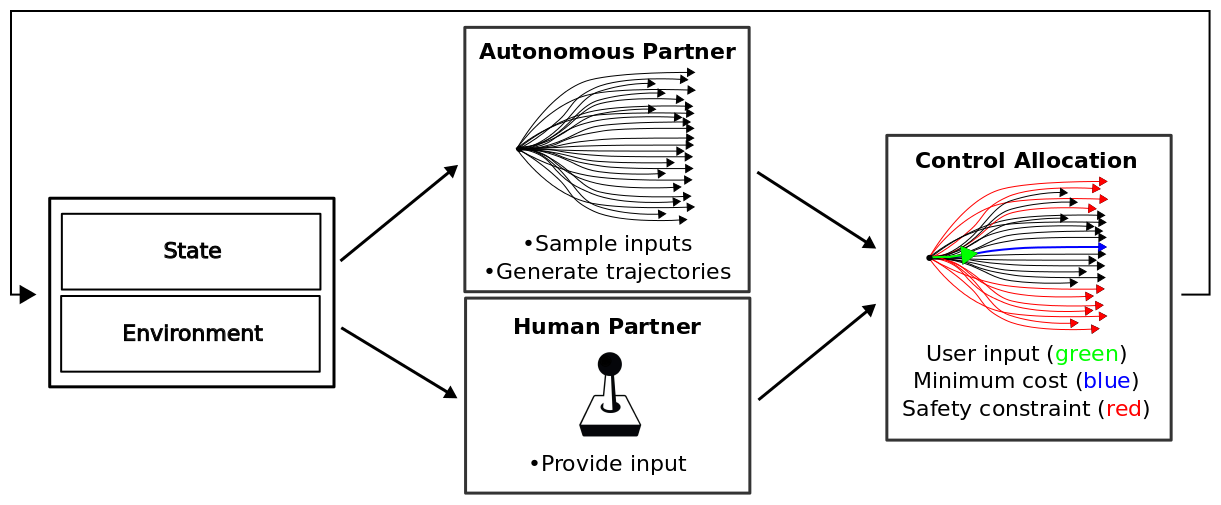}
  \caption{Pictorial representation of our model predictive minimal intervention shared control (MPMI-SC) paradigm. The autonomous partner samples densely from the input space and generates a cloud of potential trajectories using a massively parallel processor. The human partner provides their desired input.  The optimal control solution is then computed according to the Minimal Intervention Principle (MIP) and constrained based on the safety of the system over a receding horizon.}
  \label{fig-shared-control}
  \vspace{-0.4cm}
\end{figure*}

\section{Safe Minimal Intervention Shared Control}
\label{sec-approach}

In this section, we describe a theoretically correct (though computationally infeasible) solution to the problem of safe minimal intervention shared control.  We begin by defining the shared control problem mathematically.  This problem can be posed in standard optimal control terms with an additional constraint to incorporate information from both partners.  That is, our goal is to find an optimal action sequence $\bf{\bar{u}}$, and corresponding state sequence $\bf{\bar{x}}$, that minimizes the cost
\begin{equation}
\begin{aligned}
& \underset{J}{\text{minimize}}
& J(x(t), u(t)) &= \int\limits_{t=0}^{T} l(x(t), u(t)) + l_{T}(x(t)) \\
& \text{subject to}
& \dot{x}(t) &= f(x(t), u(t)),\\
& & u(t) &= g(u_h(t), u_a(t)) \\
\end{aligned}
\label{eqn-mpc}
\end{equation}

\noindent where $x(t)$ and $u(t)$ are the state and control trajectories, and $l$ and $l_{T}$ are the running and terminal costs.  The optimization is subject to constraints $f$ representing the nonlinear system dynamics, and $g$ representing the control allocation between the human ($u_h$) and autonomous partners ($u_a$).  

\subsection{Safe Control and Inevitable Collision States}

The primary constraint in $g$ relates to the safety of the human-machine system.  That is, the autonomous partner should only produce trajectories that remain safe over a receding horizon.  This requires ensuring that the system does not enter an Inevitable Collision State (ICS)~\cite{bautin2010inevitable, fraichard2004inevitable}.  ICSs refer to configurations from which it is impossible to safely recover, regardless of the control trajectory taken.  So long as the system does not enter an ICS, it is possible to develop a control strategy that results in continued safe interaction. 

\subsection{Minimum Intervention Principle}

The second feature we embed in $g$ is known as the minimal intervention principle (MIP) which states that ``an autonomous partner should only augment the human partner's control by the minimum amount necessary to achieve the desired result''~\cite{anderson2012constraint, leung2018infusing, schwarting2017parallel}.  By developing a shared control algorithm that adheres to the MIP we maximize the influence of the human partner's control at each given moment.

\subsection{Minimal Intervention Shared Control}

\begin{algorithm}[t]
  \caption{Safe Minimal Intervention Shared Control\label{alg-emisc}}
  \begin{algorithmic}[1]
  \Procedure{SMI-SC}{$x_t, u_h$}
    \State $\Gamma \gets$ all possible trajectories ($\gamma$) from $x(t)$
    \State $\Gamma_{safe} \gets \gamma \in \Gamma \: \textnormal{where} \: \gamma \notin ICS$
      \State $u_r \gets \argmin (u_h, cost(\Gamma_{safe}))$
      \State \textbf{return} $u_r$
    \EndProcedure
  \end{algorithmic}
\end{algorithm}

The described solution to minimal intervention shared control is outlined in Algorithm~\ref{alg-emisc}.  Here $\Gamma$ is the set of all possible trajectories $\gamma$ that stem from the current state $x(t)$, and $\Gamma_{safe}$ is the subset of safe trajectories.  The inputs $u_r$ and $u_h$ are the signal sent to the robot, and the signal provide by the human partner, respectively.  The $cost$ function describes how desirable a trajectory is based on it's distance from the human operator's input.  If the human's command does not lead to an ICS, their exact input will be passed to the system; otherwise, a perturbation (computed as the minimum deviation required to ensure safety) will be applied to the control signal.  

This exact (full information) algorithm is computationally infeasible to compute online for any reasonably complex human-in-the-loop system.  In particular, the solution requires exploring an infinite set of potential actions over a receding horizon to (1) ensure collision-free trajectories~\cite{wang2017safety}, and (2) select the input that most closely matches the human's desired action.  In this work, we instead propose an approximation to this solution that can be computed in real-time.

\section{Model Predictive Minimal Intervention Shared Control}
\label{sec-algorithmi-description}

To compute an approximation to the optimal solution described in the prior section, we make a few key methodological choices.  First, instead of considering \textit{all} possible trajectories from the current sate, we only consider a representative set that we generate by sampling densely (from a uniform prior) over the input space and only at the present time.  We then predict the motion of the system over a receding horizon and reject any inputs that generate a trajectory that violate the defined safety constraints.  From the subset of inputs that do not produce unsafe trajectories, we select the control signal that is closest to the human partner's input.  Notably, this approach does not require a model of the user's actions, or knowledge of a desired goal, as the user is free to dynamically adjust their objective at each timestep.  We describe each step of this algorithm in detail in the following subsections.

\subsection{Model Representation and Data-driven Approximations}
\label{sec-sub-model-rep}

The majority of related work requires hand-written (and potentially complex) models of the system and control dynamics~\cite{anderson2012constraint, leung2018infusing, schwarting2017parallel}.  In this work, we instead use a representation that is simple (i.e., linear) regardless of the underlying dynamics and can be learned from data when the model is not known \textit{a priori}.  This representation is known as the Koopman operator~\cite{koopman1931hamiltonian} and it can be used to model nonlinear dynamical systems as a linear operator because it \textit{maps functions of state to functions of state} instead operating in the original state space.  This representation is valid both when we know an analytical model of the system in the standard state space~\cite{koopman1931hamiltonian}, and when we must learn the model from data~\cite{williams2015data}.  

In this work, we opt to learn the system model from data to demonstrate the efficacy of our approach when we have no prior knowledge of the robotic system.  In particular, we use sparsity-promoting Dynamic Mode Decomposition~\cite{jovanovic2014sparsity} to select an appropriate basis and approximate the Koopman operator~\cite{koopman1931hamiltonian}.  Data-driven Koopman operators have recently been explored as a method of generating model-based control in robotics~\cite{abraham2017model} and in shared control~\cite{broad2017learning, broad2018operation}.  Unlike in prior work, however, the choice of the Koopman operator representation is explicitly motivated by our sampling-based policy generation method (see Section~\ref{sub-sec-sampling-based-control-and-safety}) and modern computational resources (e.g., multi-core CPUs and GPUs).  That is, the Koopman operator representation is particularly well suited for sampling-based control as forward predicting the state of the system requires only a single matrix-vector multiplication, and generating a large set of potential trajectories requires only a single matrix-matrix multiplication.  Both of these operations can be easily parallelized on a GPU~\cite{volkov2008benchmarking}.  Related work in model-based control for dynamic systems has utilized linear representations (e.g., Bayesian linear regression~\cite{williams2016aggressive}), however, to the best of our knowledge, ours is the first work to develop a model-based controller the integrates a Koopman operator representation with sampling-based optimal control.

\subsection{Sampling-based Optimal Control and Predictive Safety}
\label{sub-sec-sampling-based-control-and-safety}

To generate a set of \textit{unconstrained} potential trajectories the user may wish to execute, we sample $N$ inputs from an equally-spaced discretization of the control space.  By relying on a uniform prior, we make no assumptions about the user's desired action at the next step (i.e., \textit{there is no model of the user}).  These samples can also be generated stochastically.  However, stochasticity has known downsides in human-in-the-loop systems.  For example, inputs that generate motion directly along a single dimension---a common desire of human operators---are unlikely to be sampled as they exist in segments of the input space that have near zero probability mass when sampling from a continuous distribution.   

To ensure that our shared control algorithm only considers trajectories that satisfy the safety constraint described in Section~\ref{sec-approach}, we evaluate the configuration of the system at each time step over the receding horizon in each predicted trajectory.  If the system violates geometric safety checks that are defined with respect to the environment, we reject the input that generated that trajectory.  If, however, the system is safe over the entire course of the predicted trajectory, we consider that input as a viable solution.  This can be seen as a \textit{predictive notion of safety} as we evaluate the likelihood of a particular signal leading to a catastrophic failure by observing how we expect the controlled device to evolve over time.

\subsection{Model Predictive Minimal Intervention Shared Control}
\label{sub-sec-algorithm-description}

\begin{algorithm}[t]
  \caption{MPMI-SC\label{alg-mppi-sc}}
  \begin{algorithmic}[1]
  \Procedure{MPMI-SC}{$t, x_t, u_h$}
      \State $\xi \sim \mathbb{U}^{MxN}$ \Comment{unbiased control samples}
      \For {i in N \textbf{in parallel}}  \Comment{forward predict system}
        \State $t_p, x_p, safe \gets t, x_t, \textnormal{True}$
        \While{$t_p < t + T$ and $safe$} \Comment{prediction}
          \State $x_p \gets f(x_p, \xi(i)) + z_i$ \Comment{$z_i$ is Gaussian noise}
          \State $safe = \textnormal{isSafe}(x_p)$ \Comment{system safe at $x_p$}
          \State $t_p = t_p + \Delta t$ \Comment{$\Delta t$ is the timestep}
        \EndWhile\label{euclidendwhile}
        \If{$safe = \textnormal{True}$} \Comment{safe over full trajectory}
        	\State store $\gets \xi(i)$
        \EndIf
      \EndFor
      \State $u_r \gets \argmin (cost(\xi(i), u_h)) \: \forall \: \xi(i) \in \textnormal{store}$
      \State \textbf{return} $u_r$ \Comment{signal is safe and adheres to MIP}
    \EndProcedure
  \end{algorithmic}
\end{algorithm}

The full Model Predictive Minimal Intervention Shared Control (MPMI-SC) approach can be seen in Fig.~\ref{fig-shared-control} and is outlined in Algorithm~\ref{alg-mppi-sc}. The inputs are the current time $t$, the current state $x_t$ and the human partner's input $u_h$.  $\xi$ is the sampled control, $\mathbb{U}$ is the distribution the control is sampled from, $M$ is the dimensionality of the input space, $N$ is the number of samples, and $T$ is the prediction horizon.  During forward prediction (Line 6), $z_i$ is sampled i.i.d. from a white noise Gaussian process to account for inaccuracies in the dynamics model.  Notably, this computation is done \textit{in parallel on a GPU}. The learned Koopman operator system model $f(x,u)$ predicts the state $x_p$ at the next timestep $t_p$, which is evaluated for safety.  The $cost$ function minimizes the influence of the autonomous partner, and $u_r$ is the control signal sent to the robot.

\subsection{Minimal Intervention Principle and Expected Deviation}
\label{sub-sec-mip-algorithm-description}

An additional benefit of our sampling-based approach is that we can provide an explicit bound on the sub-optimality of the applied control \textit{with respect to the user's instantaneous desires}.  In particular, the deviation between the user's input and the applied control signal (\textit{when the user input is safe}) is upper bounded by half the distance between the sampled inputs.  This can be computed based on the number of samples generated at each timestep and the Lebesgue measure~\cite{tao2011introduction} of the control space.  This relationship is described in Equation~\eqref{eqn-exp-deviation}.

\begin{equation}
\mathbb{E}[\lVert u_h, u_r \rVert] = \frac{\lambda^*(U)}{2N}
\label{eqn-exp-deviation}
\end{equation}

\noindent where $u_h$ is the human partners input, $u_r$ is the signal sent to the robot, and $\lVert u, v \rVert$ is the Euclidean distance.  $\lambda^*(U)$ is the n-dimensional volume (i.e., Lebesgue measure) of the bounded input space and $N$ is the number of samples.  As the number of samples grows ($N \rightarrow \infty$), the maximum deviation between the user's input and the applied signal shrinks to 0.  However, we also note that while the number of samples (denominator) grows linearly, the Lebesgue measure (numerator) grows exponentially with each additional control dimension (i.e., the measure is defined as the Cartesian product of the intervals of each dimension).  This describes a potential issue in the scalability of our sampling based solution; however, this issue is mitigated in the majority (if not all) of our application domains as the dimensionality of the input generally remains low as the human operator must capable of providing input to the system (e.g., using a joystick).  The interval of each control dimension also generally remains small (e.g., [-1, 1] or [0, 1]) so that it is understandable by the human partner. A more in-depth discussion of the scalability of our algorithm is presented in the supplementary material.

\section{Experimental Evaluation}
\label{sec-experimental-evaluation}

We validate MPMI-SC with a human subjects study in two simulated environments which we describe below.

\subsection{Simulated Environments}
\label{sub-sec-simulated-environment}

\begin{figure}[!h]
  \begin{subfigure}[t]{.25\textwidth}
    \includegraphics[width=0.9\linewidth]{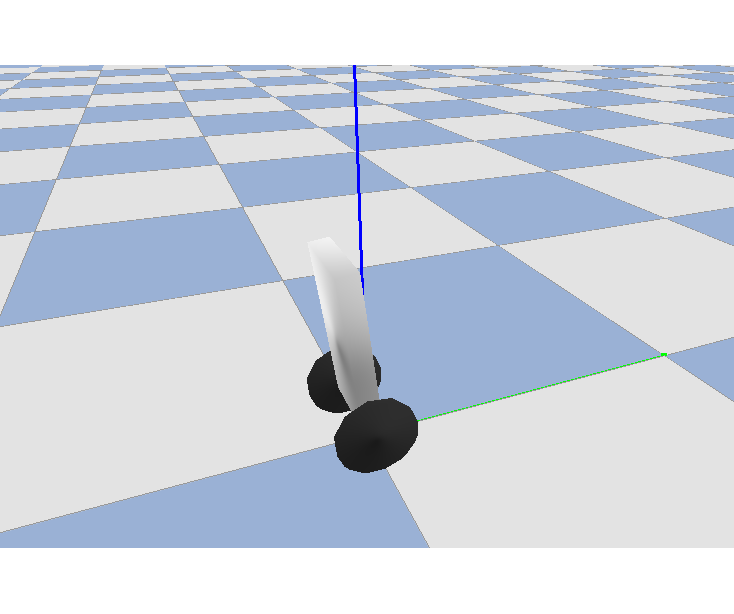}
    \caption{Simulated balance bot.}
    \label{fig-env-sim-balance-bot}
  \end{subfigure}%
  \begin{subfigure}[t]{.25\textwidth}
    \includegraphics[width=0.9\linewidth]{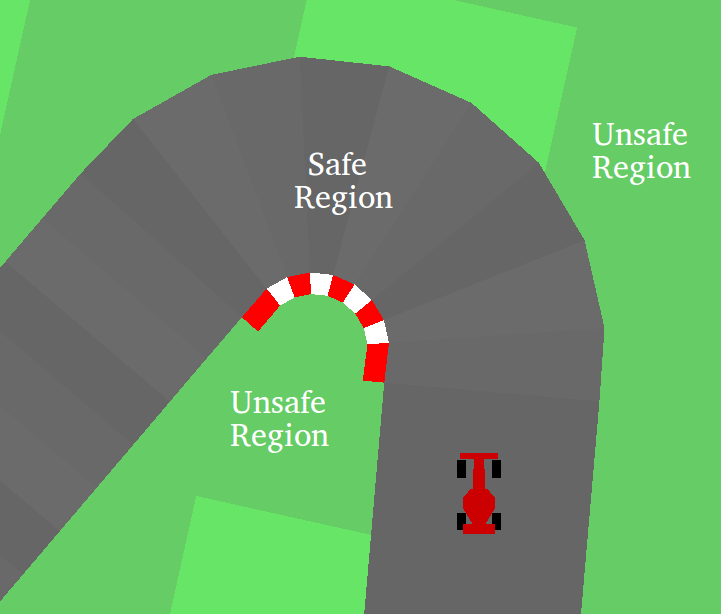}
    \caption{Simulated race car.}
    \label{fig-env-sim-race-car}
  \end{subfigure}%
  \caption{Pictorial representation of simulated environments.}
  \label{fig-env}
  \vspace{-0.3cm}
\end{figure}

The first dynamic system that users operate is a simulated balance bot (Fig.~\ref{fig-env-sim-balance-bot}).  When controlling this system, users are told (1) that they are free to move around the environment however they would like, and (2) that they should try to make sure that body of the robot does not collide with the ground (i.e., the safety constraint).  This can be challenging for a novice operator due to the stabilization requirements.  The system is based on an open-source package~\cite{chatzikonstantinou2018balance} developed using the PyBullet physics engine.  The observation space is a three-dimensional continuous vector that includes the angular position and velocity of the robot's body, and the linear velocity of the system.  The input is a one dimensional continuous signal that sets a target velocity for both wheels.    

The second dynamic system that users operate is a simulated race car (Fig.~\ref{fig-env-sim-race-car}).  When controlling this system, users are told (1) that they are free to move around the environment however they would like, and (2) that they should take caution not to drive off of the road (i.e the safety constraint).  This can be challenging for a novice operator due to the narrowness of the road and the fact that the car can go unstable (e.g., skid out) if the force applied to the system exceeds the friction limit of the ground surface.  This system is based on an open-source package released by OpenAI~\cite{brockman2016gym} and developed using the Box2D physics engine.   The observation space is a six dimensional continuous vector that includes the $(x,y,\theta)$ position of the car and the associated velocities $(\dot{x},\dot{y},\dot{\theta})$.  The input to the system is a three dimensional continuous vector that defines the desired heading for the robot, the positive acceleration (gas) and negative acceleration (break).

The complexity of these two systems makes them useful as testbeds to validate the impact of our shared control algorithm.  In particular, stabilization and environmental constraints are important challenges for robotic systems in human-centered domains where the human partner is often co-located with the mechanical system.  

\subsection{Experimental Design}
\label{sub-sec-study-protocl}

To evaluate the impact of our shared control algorithm we ran a human-subjects study (n=20) in which participants operated the system under two distinct interaction paradigms:

\begin{itemize}
\item User-only control (No assistance)
\item Model Predictive Minimal Intervention Shared Control
\end{itemize}
 
The order in which the participants saw the two paradigms was randomized and counter-balanced to account for ordering effects.  In each trial, participants were told to perform whatever action they would like so long as the system remained safe.  They were also told that in some conditions an autonomous partner would help maintain safety, but they were not told \textit{how} it would help.  A trial would end when the system violated one of the defined safety constraints or after a maximum alloted time (balance bot: 20 seconds, race car: 30 seconds). Each participant interacted with the system 10 times in each environment under each control condition.  In the race car environment, the morphology of the road was generated randomly at the start of each trial, however, the same random seeds were maintained across participants to ensure that each subject saw the same road configurations.  

\subsection{Implementation Details}

\subsubsection{Safety Computation}

In both experimental environments, the system is considered unsafe when it violates physical barriers.  For the balance bot this is defined as the robot body colliding with the ground.  For the race car this is defined as driving off the track.  To compute the safety of a predicted trajectory we evaluate the configuration of the system at each discrete timestep.  Theoretically, these geometric checks provide strong safety guarantees as we always pick the applied control signal from the subset of sampled inputs that do not violate the defined bounds.  In practice, the accuracy of this method relies both on precise system models and our ability to account for noise in the dynamics and/or sensors.  In this work, we account for these errors by adding an inflated barrier beyond the natural collision points to reduce the impact of inaccurate predictions.  Importantly, even with errors in the dynamics model, and noise in the sensor measurements, one can define an inflation radius that provides strong guarantees on the safety of the system~\cite{kousik2017safe, kousik2018bridging}.  However, in this initial work, we simply rely on a hand-tuned inflation radius.

\subsubsection{Algorithm Parameters}
\label{sub-sub-sec-parameters}

The implementation of our algorithm requires defining a small set of parameters, the two most important of which are the number of control samples $N$, and the length of the receding horizon $T$.  As these parameters increase the approximation to the true solution improves, however, this comes at a cost of increased computational complexity.  To address this issue, we provide a highly parallelized implementation of our algorithm that evaluates each trajectory independently on an NVIDIA GeForce 860M GPU (details in supplementary material).  To provide a good user experience and demonstrate the scalability of our algorithm, we select a large $N$: 10,000 for the balance bot and 10,120\footnote{Chosen to produce equally spaced samples in all 3 input dimensions.} for the race car.  The receding horizon $T$ ($T=30$ for the balance bot, $T=25$ for the race car) was chosen based on observations of each system under MPMI-SC with random inputs.

\subsubsection{Basis Selection}
Koopman operator system identification requires defining a basis to approximate the nominally infinite dimensional Hilbert space that the Koopman operator acts on.  We select this basis through data-driven methods (as described in Section~\ref{sec-sub-model-rep}).  In particular, we generate a set of 50 and 150 random basis functions for the balance bot and race car environments, respectively.  The chosen learning algorithm~\cite{jovanovic2014sparsity} selects the most relevant basis functions for modeling each system, resulting in 6 and 26 basis functions for the balance bot and race car environments, respectively.  All parameter choices are well documented in the open-source code.  

\section{Results}
\label{sec-results}

We first explore the impact of our algorithm on the safety of the human-machine system, and the users' response to the intervention of the autonomy.  We then detail the computational performance of our system using the defined parameter settings.  Finally, we provide a secondary analysis of metrics that relate to the human operator's control skill and style.

\begin{figure}[!b]
  \centering
  \includegraphics[width=\hsize]{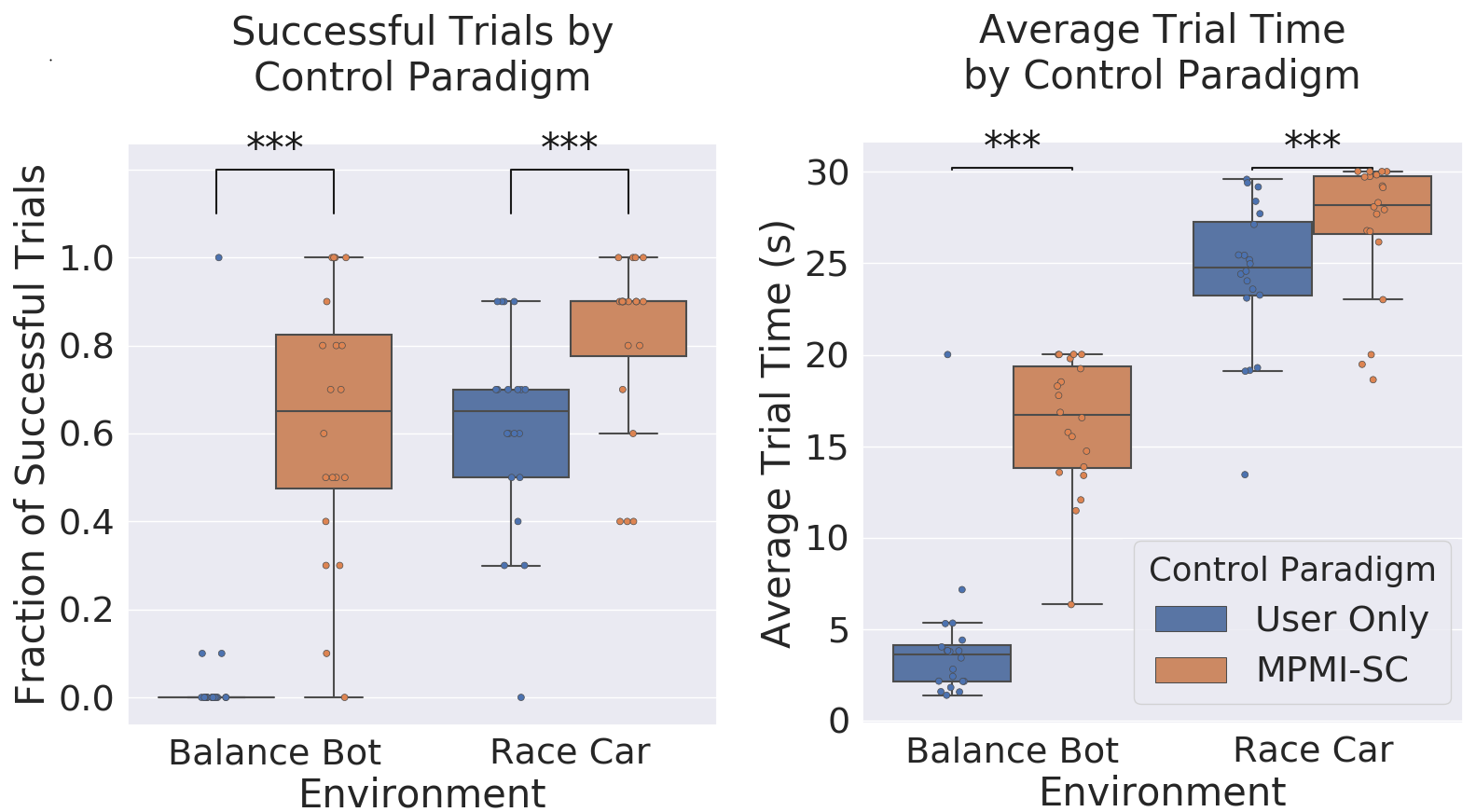}
  \caption{Average success rate (left) and average time to failure (right), broken down by environment and control paradigm. The maximum interaction time was 20s in the balance bot and 30s in the race car.  Both metrics improve significantly ($***: p < 0.005$) under shared control.}
  \label{fig-res-success}
\end{figure}

\begin{figure*}[!t]
  \centering
  \includegraphics[width=\hsize]{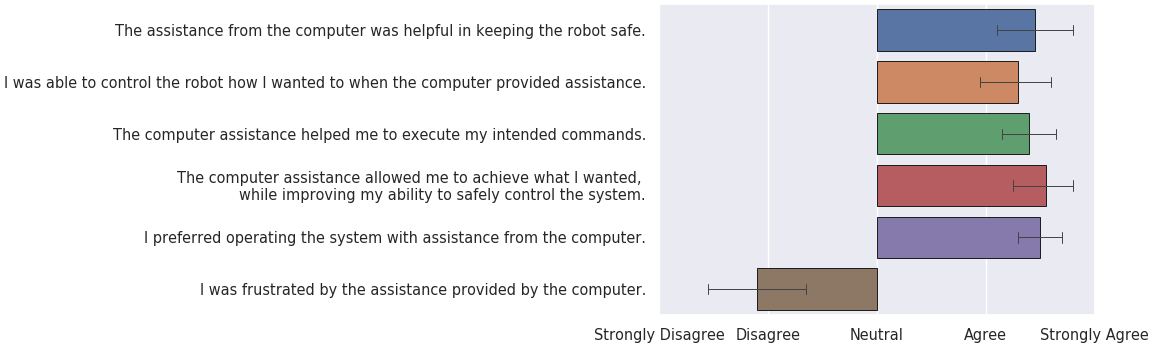}
  \caption{Average user response (agreement) to post-experiment questionnaire.  Black bars are standard deviations.}
  \label{fig-res-questionnaire}
  \vspace{-0.3cm}
\end{figure*}

\subsection{Impact of Shared Control on Safety}

To evaluate the impact of MPMI-SC on the safety of the joint system we compare (1) the average fraction of safe interactions to unsafe interactions and (2) the average time it took for the system to enter an unsafe state under each control paradigm (Fig.~\ref{fig-res-success}). To compare these values, we use a non-parametric Wilcoxon signed-rank test.  In both experimental environments we find that MPMI-SC significantly improves the rate at which users are able to safely control the system when compared to a user-only control paradigm ($p < 0.005$).  Similarly, we find that users are able to safely control the system for a significantly longer amount of time under shared control then when under user-only control ($p < 0.005$). 

\subsection{User Acceptance of Shared Control Paradigm}

We next evaluate the users' acceptance of the assistance provided by the autonomous system.  As mentioned in Section~\ref{sec-background-and-related-work}, the majority of shared control systems assist a user in achieving a specific task.  The assistance can therefore come at the expense of user satisfaction as the human partner often feels that they are fighting the autonomy~\cite{erdogan2017prediction}.  In contrast, we develop a task-agnostic shared control paradigm that adheres to the human partner's desires at each instant.  

To evaluate the user experience under MPMI-SC we asked participants to fill out a post-experiment questionnaire (Fig.~\ref{fig-res-questionnaire}).  Statements were rated on a 5-point Likert-type scale where 1 represents strong disagreement and 5 represents strong agreement.  Overall, user's felt the assistance provided by MPMI-SC helped them keep the robot safe and execute their indented commands.  Perhaps most telling is that the participants strongly preferred operating the system with assistance from the computer, and did not feel frustrated by the assistance.

\subsection{System Performance}
\label{sub-sec-system-performance}

We now describe the computational efficiency of MPMI-SC using the parameters defined in Section~\ref{sub-sub-sec-parameters}. Our algorithm is capable of generating trajectories at $\sim$7000 Hz and $\sim$3500 Hz in the balance bot and race car environments.  Incorporating the safety checks, the system runs at $\sim$100 Hz and $\sim$60 Hz, respectively (i.e., between 600,000 and 1,000,000 trajectories every second).  MPMI-SC is faster in the balance bot as it relies on a smaller basis and more efficient safety checks.  A detailed description of our GPU implementation and the scalability of MPMI-SC is provided in the supplementary material.

We also compute the maximum possible deviation between a user's input and the closest safe signal at each timestep.  This value is determined by evaluating Equation~\eqref{eqn-exp-deviation} with a known n-dimensional input volume ($\lambda*$) and number of samples ($N$).  If the user's input is considered safe over the receding horizon (i.e., through our prediction method described in Section~\ref{sub-sec-sampling-based-control-and-safety}), this value represents \textit{the maximum influence of the autonomous partner on the applied signal}.  If the user's input is not safe over the predicted trajectory, this bound represents \textit{the maximum possible difference between the true closest safe signal and the applied signal}.  For the balance bot, the maximum deviation is $\pmb{\mathbb{E} = 0.0001}$ because $\lambda^*(U)$ = 2 (i.e., 1D input that spans $[-1, 1]$) and $N = 10,000$.  For the race car, the maximum deviation is $\pmb{\mathbb{E} = 0.0002}$ because $\lambda^*(U)$ = 4 (i.e., 3D input where the heading spans $[-1,1]$, the gas spans $[0,1]$, and the break spans $[-1, 0]$) and $N = 10,120$.  

How large an impact an intervention of the described magnitudes has on the dynamics of a dynamic system is dependent on each particular machine.  However, we note that both values described in this work are \textit{very} small.  We therefore hypothesize that there is likely no discernible difference between the effect of the intended signal and the applied signal when the user's input is considered safe.  If there were a large impact on the system dynamics due to an intervention of this magnitude, the system may be chaotic and therefore challenging to operate no matter what controller was used~\cite{fradkov2005control}.

\subsection{User Control Skill and Style}

Finally, we provide a secondary analysis of the data collected during our experiment to evaluate the users' control skill and style.  The proposed metrics are easy to compute based on the users' input and the trajectories generated at each timestep.

\subsubsection{Operator Control Skill}
\label{sub-sub-sec-operator-control-skill}

We first calculate the average observed deviation between the user input and the \textit{closest safe signal} as a proxy for the users' understanding of the system dynamics and their control skill.  To illustrate this relationship consider a car making a tight turn.  If the human partner understands that they need to change their heading and speed to stay on the road during the turn, this metric will remain low (Sec.~\ref{sub-sec-mip-algorithm-description}).  If, however, the human partner does not understand this relationship and relies on the autonomous partner to maintain the safety of the system during the turn, this metric will increase.  The average value of this metric will be within the described bound ($\pmb{\mathbb{E}}$, above) if, and only if, \textit{all} of the user's controls are considered safe during the course of their interaction.  Otherwise this value will increase according to the minimal amount required to maintain safety.  

\subsubsection{Operator Control Style}
\label{sub-sub-sec-riskiness-in-control}

We then calculate the average percentage of sampled rollouts that are safe at each timestep as a proxy for the user's control style.  A lower percentage of safe trajectories suggests that the user is operating the system in a more \textit{dangerous} manner (i.e., the system may be closer to entering an ICS).  However, we note that this metric alone does not directly relate to the user's control skill as there are many cases in which one may trade off notions of safety for performance.  To illustrate this relationship consider, again, a person in a car taking a tight turn to decrease the time of their drive.  If this value correlates positively with safe control of the system, we can say that the operator likely has a high degree of skill and is explicitly trading off conventional notions of safety (e.g., distance from an obstacle) for improved performance.  However, if this metric correlates positively with unsafe behavior, it is likely that the operator is unskilled and making poor control decisions. 

\begin{table}[!t]
\centering
\begin{tabular}{|c|c|c|c|}
\hline
\textbf{Metric} & \textbf{Control} & \textbf{Balance Bot} & \textbf{Race Car} \\
\hline
\multirow{ 2}{*}{\textbf{Avg. Deviation}} & User-only & 0.46 $\pm$ 0.37 & 0.05 $\pm$ 0.03 \\
 & MPMI-SC & 0.21 $\pm$ 0.21 & 0.07 $\pm$ 0.06 \\
\hline
\multirow{ 2}{*}{\textbf{Avg. \% Safe Rollouts}} & User-only & 0.28 $\pm$ 0.15 & 0.38 $\pm$ 0.09  \\
 & MPMI-SC & 0.50 $\pm$ 0.10 & 0.35 $\pm$ 0.08 \\
\hline
\end{tabular}
\vspace{0.1cm}
\caption{Average deviation and average percentage safe rollouts, broken down by environment and by control condition.}
\label{table-metrics}
\vspace{-0.5cm}
\end{table}

\subsubsection{Analysis}
 
In the balance bot environment, we find that the user's input more closely aligns with the closest safe signal under MPMI-SC (Table~\ref{table-metrics}).  We also observe that the system remains in a state where more potential actions remain safe over the horizon.  We interpret these results as evidence that the users provided more competent control with assistance than without.  In the race car environment, we find that, in both control conditions, the user's input is nearly equally aligned with the closest safe signal, and that there are similar percentages of potentially safe actions.  We interpret these results as evidence that MPMI-SC had less impact in this environment then in the balance bot.  Evidence of this interpretation can also be seen in Figure~\ref{fig-res-success} where we find larger raw differences between the safety metrics in the balance bot environment than in the race car environment.  However, the differences between each of these primary safety metrics are statistically significant suggesting that, even when MPMI-SC is less impactful, it still meaningfully improves the safety of the joint system.  Both proposed metrics are indirect measures of the user's control skill and style, but the preliminary results (Table~\ref{table-metrics}) suggest they warrant further investigation.  In future work we plan to evaluate how they evolve over time for an individual operator.

\section{Discussion}
\label{sec-discussion}

\subsection{Study Observations}

One piece of information that is not reflected in our analysis is how people alter their \textit{control strategy} when they are under different paradigms. For example, participants were observed \textit{testing} the limits of the assistance.  That is, users would intentionally operate the system at the boundary of safety, and even act in an adversarial manner to test the reliability of MPMI-SC.  In contrast, under user-only control, participants were much more cautious.  This is potentially a consequence of not explaining \textit{how} the autonomous partner would provide assistance; however, we also believe this behavior aligns with human nature as people often \textit{explore} while they learn. 

Another observation relates to why MPPI-SC had a larger impact on the balance bot than on the race car.  In particular, participants were more likely to have prior experience operating car-like systems then unstable machines like the balance bot, which could mean they were better able to provide safe control based solely on experience and intuition.

\subsection{Limitation in Prediction of Safety}

As described in Section~\ref{sec-algorithmi-description}, the control signal sent to the robot is selected based on the user's desires at each instant and therefore does not require a model of the user.  However, to compute the safety of a given control action, we implicitly embed a na\"ive model of the user that assumes the human partner will continue to apply nearly the same input over the time-horizon.  The negative implication of this assumption is that we will occasionally reject user inputs that \textit{seem} dangerous, but are not in reality if the user quickly adjusts their strategy.  Therefore, there will be trajectories that the person would like to execute and are safe (by recovering at a later timestep) that MPMI-SC incorrectly rejects.  Despite this limitation, it is possible that our iterative, receding horizon approach alleviates the impact on the user's acceptance by recomputing the set of dangerous actions at each timestep.  We leave a deeper evaluation of this idea to future work.

\section{Conclusion}
\label{sec-conclusion}

In conclusion, we described a shared control paradigm that enhances a human partner's ability to operate complex, dynamic machines by incorporating safety constraints without explicit knowledge of the user's long-term objective.  Our approach relies on a simple representation of the joint system (i.e., the Koopman operator) which, in turn, means we can very quickly generate and evaluate the safety of a large number of potential trajectories through the parallelization capabilities of a GPU.  Importantly, this representation can be learned from data and therefore generalizes to any pair of partners.  Finally, our approach adheres to the minimal intervention principle to ensure that the human partner is allocated the majority of the decision making authority throughout the interaction. 

We evaluated the efficacy of our Model Predictive Minimal Intervention Shared Control (MPMI-SC) paradigm with a human-subjects study consisting of 20 participants.  The results demonstrated that our approach is able to improve the general safety of the joint system without \textit{a priori} knowledge of the user's desires.  Additionally, we found that the participants enjoyed the assistance provided by the autonomy (and reported low levels of frustration), a feature lacking in many shared control paradigms~\cite{erdogan2017prediction}.  We have released our code with an open-source license on GitHub at \url{https://github.com/asbroad/mpmi_shared_control}.

\section*{Acknowledgments}
This material is based upon work supported by the National Science Foundation under Grants CNS 1329891 \& 1837515. Any opinions, findings and conclusions or recommendations expressed in this material are those of the authors and do not necessarily reflect the views of the aforementioned institutions.  The authors would also like to thank Emily Bernstein for her help in the study design.

\bibliographystyle{plainnat}
\bibliography{references}

\begin{thebibliography}{37}
\providecommand{\natexlab}[1]{#1}
\providecommand{\url}[1]{\texttt{#1}}
\expandafter\ifx\csname urlstyle\endcsname\relax
  \providecommand{\doi}[1]{doi: #1}\else
  \providecommand{\doi}{doi: \begingroup \urlstyle{rm}\Url}\fi

\bibitem[Abbink et~al.(2018)Abbink, Carlson, Mulder, de~Winter, Aminravan,
  Gibo, and Boer]{abbink2018topology}
David~A Abbink, Tom Carlson, Mark Mulder, Joost~CF de~Winter, Farzad Aminravan,
  Tricia~L Gibo, and Erwin~R Boer.
\newblock \href{https://ieeexplore.ieee.org/abstract/document/8353134/}{A
  Topology of Shared Control Systems---Finding Common Ground in Diversity}.
\newblock \emph{IEEE Transactions on Human-Machine Systems}, \penalty0
  (99):\penalty0 1--17, 2018.

\bibitem[Abraham et~al.(2017)Abraham, Torre, and Murphey]{abraham2017model}
Ian Abraham, Gerardo De~La Torre, and Todd~D Murphey.
\newblock \href{http://www.roboticsproceedings.org/rss13/p52.pdf}{Model-Based
  Control Using Koopman Operators}.
\newblock In \emph{Robotics: Science and Systems}, 2017.

\bibitem[Anderson et~al.(2012)Anderson, Karumanchi, and
  Iagnemma]{anderson2012constraint}
Sterling~J Anderson, Sisir~B Karumanchi, and Karl Iagnemma.
\newblock \href{https://ieeexplore.ieee.org/document/6232153/}{Constraint-based
  Planning and Control for Safe, Shared Control of Ground Vehicles}.
\newblock In \emph{IEEE International Vehicles Symposium}, 2012.

\bibitem[Anderson et~al.(2013)Anderson, Karumanchi, Iagnemma, and
  Walker]{anderson2013intelligent}
Sterling~J Anderson, Sisir~B Karumanchi, Karl Iagnemma, and James~M Walker.
\newblock \href{https://ieeexplore.ieee.org/document/6507273/}{The Intelligent
  CoPilot: A Constraint-based Approach to Shared-Adaptive Control of Ground
  Vehicles}.
\newblock \emph{Intelligent Transportation Systems Magazine}, 5\penalty0
  (2):\penalty0 45--54, 2013.

\bibitem[Anderson et~al.(2014)Anderson, Walker, and
  Iagnemma]{anderson2014experimental}
Sterling~J Anderson, James~M Walker, and Karl Iagnemma.
\newblock \href{https://ieeexplore.ieee.org/document/6766255/}{Experimental
  Performance Analysis of a Homotopy-based Shared Autonomy Framework}.
\newblock \emph{Transactions on Human-Machine Systems}, 44\penalty0
  (2):\penalty0 190--199, 2014.

\bibitem[Bautin et~al.(2010)Bautin, Martinez-Gomez, and
  Fraichard]{bautin2010inevitable}
Antoine Bautin, Luis Martinez-Gomez, and Thierry Fraichard.
\newblock
  \href{https://ieeexplore.ieee.org/abstract/document/5509233/}{Inevitable
  Collision States: A Probabilistic Perspective}.
\newblock In \emph{International Conference on Robotics and Automation}, pages
  4022--4027. IEEE, 2010.

\bibitem[Broad et~al.(2017)Broad, Murphey, and Argall]{broad2017learning}
Alexander Broad, Todd Murphey, and Brenna Argall.
\newblock \href{http://www.roboticsproceedings.org/rss13/p37.pdf}{Learning
  Models for Shared Control of Human-Machine Systems with Unknown Dynamics}.
\newblock In \emph{Robotics: Science and Systems}, 2017.

\bibitem[Broad et~al.(2018)Broad, Murphey, and Argall]{broad2018operation}
Alexander Broad, Todd Murphey, and Brenna Argall.
\newblock
  \href{https://nxr.northwestern.edu/publications/operation-and-imitation-under-safety}{Operation
  and Imitation under Safety-Aware Shared Control}.
\newblock In \emph{International Workshop on the Algorithmic Foundations of
  Robotics}, 2018.

\bibitem[Brockman et~al.(2016)Brockman, Cheung, Pettersson, Schneider,
  Schulman, Tang, and Zaremba]{brockman2016gym}
Greg Brockman, Vicki Cheung, Ludwig Pettersson, Jonas Schneider, John Schulman,
  Jie Tang, and Wojciech Zaremba.
\newblock \href{https://arxiv.org/abs/1606.01540}{OpenAI Gym}.
\newblock \emph{arXiv}, abs/1606.01540, 2016.

\bibitem[Carlson and Demiris()]{carlson2012collaborative}
Tom Carlson and Yiannis Demiris.
\newblock \href{https://ieeexplore.ieee.org/document/6135817/}{Collaborative
  Control for a Robotic Wheelchair: Evaluation of Performance, Attention, and
  Workload}.
\newblock \emph{Transactions on Systems, Man, and Cybernetics}, 42\penalty0
  (3):\penalty0 876--888.

\bibitem[Carlson and Demiris(2008)]{carlson2008human}
Tom Carlson and Yiannis Demiris.
\newblock \href{https://ieeexplore.ieee.org/document/4543814/}{Human-Wheelchair
  Collaboration through Prediction of Intention and Adaptive Assistance}.
\newblock In \emph{IEEE International Conference on Robotics and Automation},
  pages 3926--3931, 2008.

\bibitem[Chatzikonstantinou(2018)]{chatzikonstantinou2018balance}
Yannis Chatzikonstantinou.
\newblock Balance bot.
\newblock \url{https://github.com/yconst/balance-bot}, 2018.

\bibitem[Dragan and Srinivasa(2013)]{dragan2013policy}
Anca~D Dragan and Siddhartha~S Srinivasa.
\newblock \href{http://journals.sagepub.com/doi/abs/10.1177/0278364913490324}{A
  Policy-Blending Formalism for Shared Control}.
\newblock \emph{The International Journal of Robotics Research}, 32\penalty0
  (7):\penalty0 790--805, 2013.

\bibitem[Erdogan and Argall(2017)]{erdogan2017prediction}
Ahmetcan Erdogan and Brenna~D Argall.
\newblock \href{https://ieeexplore.ieee.org/document/8009397/}{Prediction of
  User Preference over Shared Control Paradigms for a Robotic Wheelchair}.
\newblock In \emph{IEEE International Conference on Rehabilitation Robotics},
  July 2017.

\bibitem[Erlien et~al.(2016)Erlien, Fujita, and Gerdes]{erlien2016shared}
Stephen~M Erlien, Susumu Fujita, and Joseph~Christian Gerdes.
\newblock \href{https://ieeexplore.ieee.org/document/7317796/}{Shared Steering
  Control using Safe Envelopes for Obstacle Avoidance and Vehicle Stability}.
\newblock \emph{Transactions on Intelligent Transportation Systems},
  17\penalty0 (2):\penalty0 441--451, 2016.

\bibitem[Fradkov and Evans(2005)]{fradkov2005control}
Alexander~L Fradkov and Robin~J Evans.
\newblock
  \href{https://www.sciencedirect.com/science/article/pii/S1367578805000040}{Control
  of Chaos: Methods and Applications in Engineering}.
\newblock \emph{Annual Reviews in Control}, 29\penalty0 (1):\penalty0 33--56,
  2005.

\bibitem[Fraichard and Asama(2004)]{fraichard2004inevitable}
Thierry Fraichard and Hajime Asama.
\newblock \href{https://ieeexplore.ieee.org/document/1250659/}{Inevitable
  Collision States—A Step Towards Safer Robots?}
\newblock \emph{Advanced Robotics}, 18\penalty0 (10):\penalty0 1001--1024,
  2004.

\bibitem[Javdani et~al.(2018)Javdani, Admoni, Pellegrinelli, Srinivasa, and
  Bagnell]{javdani2018shared}
Shervin Javdani, Henny Admoni, Stefania Pellegrinelli, Siddhartha~S Srinivasa,
  and J~Andrew Bagnell.
\newblock
  \href{https://journals.sagepub.com/doi/full/10.1177/0278364918776060}{Shared
  Autonomy via Hindsight Optimization for Teleoperation and Teaming}.
\newblock \emph{The International Journal of Robotics Research}, 37\penalty0
  (7):\penalty0 717--742, 2018.

\bibitem[Jovanovi{\'c} et~al.(2014)Jovanovi{\'c}, Schmid, and
  Nichols]{jovanovic2014sparsity}
Mihailo~R Jovanovi{\'c}, Peter~J Schmid, and Joseph~W Nichols.
\newblock
  \href{https://aip.scitation.org/doi/abs/10.1063/1.4863670}{Sparsity-promoting
  Dynamic Mode Decomposition}.
\newblock \emph{Physics of Fluids}, 26\penalty0 (2):\penalty0 024103, 2014.

\bibitem[Kim et~al.(2006)Kim, Biggs, Schloerb, Carmena, Lebedev, Nicolelis, and
  Srinivasan]{kim2006continuous}
Hyun~K Kim, J~Biggs, W~Schloerb, M~Carmena, Mikhail~A Lebedev, Miguel~AL
  Nicolelis, and Mandayam~A Srinivasan.
\newblock
  \href{https://ieeexplore.ieee.org/abstract/document/1634510/}{Continuous
  Shared Control for Stabilizing Reaching and Grasping with Brain-Machine
  Interfaces}.
\newblock \emph{IEEE Transactions on Biomedical Engineering}, 53\penalty0
  (6):\penalty0 1164--1173, 2006.

\bibitem[Koopman(1931)]{koopman1931hamiltonian}
Bernard~O Koopman.
\newblock \href{http://www.pnas.org/content/17/5/315.short}{Hamiltonian Systems
  and Transformation in Hilbert space}.
\newblock \emph{Proceedings of the National Academy of Sciences}, 17\penalty0
  (5):\penalty0 315--318, 1931.

\bibitem[Kousik et~al.(2017)Kousik, Vaskov, Johnson-Roberson, and
  Vasudevan]{kousik2017safe}
Shreyas Kousik, Sean Vaskov, Matthew Johnson-Roberson, and Ram Vasudevan.
\newblock
  \href{http://proceedings.asmedigitalcollection.asme.org/proceeding.aspx?articleid=2663490}{Safe
  Trajectory Synthesis for Autonomous Driving in Unforeseen Environments}.
\newblock In \emph{ASME Dynamic Systems and Control Conference}, 2017.

\bibitem[Kousik et~al.(2018)Kousik, Vaskov, Bu, Johnson-Roberson, and
  Vasudevan]{kousik2018bridging}
Shreyas Kousik, Sean Vaskov, Fan Bu, Matthew Johnson-Roberson, and Ram
  Vasudevan.
\newblock \href{https://arxiv.org/abs/1809.06746}{Bridging the Gap Between
  Safety and Real-Time Performance in Receding-Horizon Trajectory Design for
  Mobile Robots}.
\newblock \emph{arXiv:1809.06746}, 2018.

\bibitem[LaValle and Kuffner~Jr(2001)]{lavalle2001randomized}
Steven~M LaValle and James~J Kuffner~Jr.
\newblock
  \href{http://journals.sagepub.com/doi/abs/10.1177/02783640122067453}{Randomized
  Kinodynamic Planning}.
\newblock \emph{The International Journal of Robotics Research}, 20\penalty0
  (5):\penalty0 378--400, 2001.

\bibitem[Leung et~al.(2018)Leung, Schmerling, Chen, Talbot, Gerdes, and
  Pavone]{leung2018infusing}
K.~Leung, E.~Schmerling, M.~Chen, J.~Talbot, J.~C. Gerdes, and M.~Pavone.
\newblock
  \href{http://asl.stanford.edu/wp-content/papercite-data/pdf/Leung.Schmerling.Chen.ea.ISER18.pdf}{On
  Infusing Reachability-Based Safety Assurance within Probabilistic Planning
  Frameworks for Human-Robot Vehicle Interactions}.
\newblock In \emph{International Symposium on Experimental Robotics}, 2018.

\bibitem[Musi{\'c} and Hirche(2017)]{music2017control}
Selma Musi{\'c} and Sandra Hirche.
\newblock
  \href{https://www.sciencedirect.com/science/article/pii/S1367578817301153}{Control
  Sharing in Human-Robot Team Interaction}.
\newblock \emph{Annual Reviews in Control}, 2017.

\bibitem[Reddy et~al.(2018)Reddy, Dragan, and Levine]{reddy2018shared}
Siddharth Reddy, Anca~D Dragan, and Sergey Levine.
\newblock \href{http://www.roboticsproceedings.org/rss14/p05.pdf}{Shared
  Autonomy via Deep Reinforcement Learning}.
\newblock In \emph{Robotics: Science and Systems}, 2018.

\bibitem[Schwarting et~al.(2017{\natexlab{a}})Schwarting, Alonso-Mora, Paull,
  Karaman, and Rus]{schwarting2017parallel}
Wilko Schwarting, Javier Alonso-Mora, Liam Paull, Sertac Karaman, and Daniela
  Rus.
\newblock \href{https://ieeexplore.ieee.org/document/7989224/}{Parallel
  autonomy in automated vehicles: Safe motion generation with minimal
  intervention}.
\newblock In \emph{IEEE International Conference on Robotics and Automation},
  2017{\natexlab{a}}.

\bibitem[Schwarting et~al.(2017{\natexlab{b}})Schwarting, Alonso-Mora, Paull,
  Karaman, and Rus]{schwarting2017safe}
Wilko Schwarting, Javier Alonso-Mora, Liam Paull, Sertac Karaman, and Daniela
  Rus.
\newblock \href{https://ieeexplore.ieee.org/document/8207782/}{Safe Nonlinear
  Trajectory Generation for Parallel Autonomy with a Dynamic Vehicle Model}.
\newblock \emph{IEEE Transactions on Intelligent Transportation Systems},
  \penalty0 (99):\penalty0 1--15, 2017{\natexlab{b}}.

\bibitem[Shia et~al.(2014)Shia, Gao, Vasudevan, Campbell, Lin, Borrelli, and
  Bajcsy]{shia2014semiautonomous}
Victor~A Shia, Yiqi Gao, Ramanarayan Vasudevan, Katherine~Driggs Campbell,
  Theresa Lin, Francesco Borrelli, and Ruzena Bajcsy.
\newblock
  \href{https://ieeexplore.ieee.org/abstract/document/6828752/}{Semiautonomous
  Vehicular Control using Driver Modeling}.
\newblock \emph{IEEE Transactions on Intelligent Transportation Systems},
  15\penalty0 (6):\penalty0 2696--2709, 2014.

\bibitem[Tao(2011)]{tao2011introduction}
Terence Tao.
\newblock \emph{\href{https://bookstore.ams.org/gsm-126}{An Introduction to
  Measure Theory}}.
\newblock American Mathematical Society, 2011.

\bibitem[Theodorou et~al.(2010)Theodorou, Buchli, and
  Schaal]{theodorou2010generalized}
Evangelos Theodorou, Jonas Buchli, and Stefan Schaal.
\newblock \href{http://www.jmlr.org/papers/v11/theodorou10a.html}{A Generalized
  Path Integral Control Approach to Reinforcement Learning}.
\newblock \emph{Journal of Machine Learning Research}, 11\penalty0
  (Nov):\penalty0 3137--3181, 2010.

\bibitem[Volkov and Demmel(2008)]{volkov2008benchmarking}
Vasily Volkov and James~W Demmel.
\newblock
  \href{https://mc.stanford.edu/cgi-bin/images/6/65/SC08_Volkov_GPU.pdf}{Benchmarking
  GPUs to Tune Dense Linear Algebra}.
\newblock In \emph{International Conference for High Performance Computing,
  Networking, Storage and Analysis}, pages 1--11. IEEE, 2008.

\bibitem[Wang et~al.(2017)Wang, Ames, and Egerstedt]{wang2017safety}
Li~Wang, Aaron~D Ames, and Magnus Egerstedt.
\newblock \href{https://ieeexplore.ieee.org/document/7857061}{Safety Barrier
  Certificates for Collisions-Free Multirobot Systems}.
\newblock \emph{IEEE Transactions on Robotics}, 33\penalty0 (3):\penalty0
  661--674, 2017.

\bibitem[Williams et~al.(2016)Williams, Drews, Goldfain, Rehg, and
  Theodorou]{williams2016aggressive}
Grady Williams, Paul Drews, Brian Goldfain, James~M Rehg, and Evangelos~A
  Theodorou.
\newblock
  \href{https://ieeexplore.ieee.org/stamp/stamp.jsp?tp=&arnumber=7487277}{Aggressive
  Driving with Model Predictive Path Integral Control}.
\newblock In \emph{International Conference on Robotics and Automation (ICRA)},
  pages 1433--1440. IEEE, 2016.

\bibitem[Williams et~al.(2017)Williams, Aldrich, and
  Theodorou]{williams2017model}
Grady Williams, Andrew Aldrich, and Evangelos~A Theodorou.
\newblock \href{https://arc.aiaa.org/doi/abs/10.2514/1.G001921}{Model
  Predictive Path Integral Control: From Theory to Parallel Computation}.
\newblock \emph{Journal of Guidance, Control, and Dynamics}, 40\penalty0
  (2):\penalty0 344--357, 2017.

\bibitem[Williams et~al.(2015)Williams, Kevrekidis, and
  Rowley]{williams2015data}
Matthew~O Williams, Ioannis~G Kevrekidis, and Clarence~W Rowley.
\newblock \href{https://link.springer.com/article/10.1007/s00332-015-9258-5}{A
  Data-driven Approximation of the Koopman Operator: Extending Dynamic Mode
  Decomposition}.
\newblock \emph{Journal of Nonlinear Science}, 25\penalty0 (6):\penalty0
  1307--1346, 2015.

\end{thebibliography}

\end{document}